\documentclass[onecolumn,journal]{IEEEtran}
\usepackage{amssymb}
\usepackage{mathrsfs}
\usepackage{amsfonts}
\usepackage{amsmath}
\usepackage{graphicx}
\usepackage{multirow}
\usepackage{caption2}
\usepackage{float}
\usepackage{booktabs}
\usepackage{algorithm}
\usepackage{algorithmic}
\usepackage{subfigure}
\usepackage{psfrag}
\usepackage{txfonts}
\usepackage{latexsym,bm}
\usepackage{color}
\usepackage[table]{xcolor}
\usepackage{url}

\newtheorem{theorem}{Theorem}

\newtheorem{lemma}{Lemma}

\newtheorem{definition}{Definition}

\ifCLASSINFOpdf
\else
\fi

\begin{document}
%
\title{Universal Consistency of Deep Convolutional Neural Networks}

\author{Shao-Bo Lin, Kaidong Wang, Yao Wang, and~Ding-Xuan Zhou
\IEEEcompsocitemizethanks{\IEEEcompsocthanksitem S. B. Lin, K. Wang and Y. Wang are  with the Center for Intelligent Decision-Making and Machine Learning, School of Management, Xi'an Jiaotong University, Xi'an 710049,   P R China.  D. X. Zhou is with
School of Data Science and Department of Mathematics, City University of Hong Kong, {  Hong Kong}. The corresponding author is Y. Wang (email:
yao.s.wang@gmail.com).}}

 \IEEEcompsoctitleabstractindextext{
\begin{abstract}
Compared with  avid research activities of deep convolutional neural networks (DCNNs) in practice, the study of theoretical behaviors of DCNNs lags heavily behind. In particular, the universal consistency of DCNNs remains open. In this paper, we prove that implementing empirical risk minimization on DCNNs with expansive convolution (with zero-padding) is strongly universally consistent. Motivated by the universal consistency, we conduct a series of experiments to show that  without any fully connected layers,  DCNNs with expansive convolution perform not worse than the widely used deep neural networks with hybrid structure containing contracting (without zero-padding) convolution layers and several fully connected layers.
\end{abstract}

}

\maketitle

\IEEEdisplaynotcompsoctitleabstractindextext

\IEEEpeerreviewmaketitle
%
%

\section{Introduction}
The great success of deep learning \cite{Goodfellow2016} in practice stimulates avid research activities to understand the magic behind it. The reasons for success can be attributed to the depth of networks \cite{Kohler2017,Schmidt2020}, massiveness of data \cite{Lin2019,Chui2020}, fast developed optimization algorithms \cite{Mei2018,Allen-Zhu2019} and more importantly,  architectures \cite{Bengio2013,He2016}  that reduce the number of free parameters of networks while maintain their excellent performances  in feature extraction and function representations.
Deep convolutional neural networks (DCNNs) that equip deep neural networks with convolutional structures are one of the most popular networks used in image processing \cite{Krizhevsky2012}, game theory  \cite{Silver2016}, signal processing \cite{Kachuee2018}, among many others. 

We are interested in DCNNs induced by one-dimensional convolution, one channel and  the rectifier linear unit (ReLU) activation function.   As in \cite{Zhou2020a}, there are not any fully connected layers in DCNNs considered in this paper.
The convolution of two functions $f$ and $h$ on $\mathbb R$ is defined by
$$
          h\otimes f(x)=\int_{-\infty}^\infty f(x')h(x-x')dx', \qquad x\in {\mathbb R}.
$$
Discretely, let $\vec{w}=(w_j)_{j=-\infty}^\infty$ be a filter of  of length $s$, i.e.  $w_j^k\neq0$ only for $0\leq j\leq s$.
Two widely used types of 1-D convolution of $\vec{w}$ with a vector $\vec{v}=(v_1,\dots,v_D)^T$, regarded as a sequence on ${\mathbb Z}$ supported in $\{1, \ldots, D\}$, are the expansive convolution (also called convolution with zero-padding) denoted by $\vec{w}* \vec{v}$ and contracting convolution (or convolution without zero-padding) $\vec{w}\star\vec{v}$ with $j$-th components
\begin{equation}\label{convolution}
    (\vec{w}*\vec{v})_j=\sum_{\ell=1}^{D}w_{j-\ell}v_j,\qquad j=1,\dots,D+s,
\end{equation}
and
\begin{equation}\label{convolution1}
    (\vec{w}\star\vec{v})_j=\sum_{\ell=s+1}^{D-s}w_{j-\ell}v_j,\qquad j=1,\dots,D-s.
\end{equation}
From \eqref{convolution} and \eqref{convolution1}, it is easy to derive \cite{Zhou2020a} that there have $D\times (D+s)$ sparse Toeplitz type matrix  $\widetilde{W}$  and $D\times(D-s)$ one $\widetilde{W'}$ such that
\begin{equation}\label{sparse-structure}
    \vec{w}* \vec{v}=\widetilde{W}\vec{v},\qquad\mbox{and}\quad \vec{w}\star \vec{v}=\widetilde{W}'\vec{v},\qquad \forall \vec{v}\in\mathbb R^D.
\end{equation}

Let $\sigma(t)=\max\{0,t\}$ be  ReLU and
$L\in\mathbb N$ be the number of hidden layers. Given a set of filters $\{\vec{w}_k\}_{k=1}^L$, a set of thresholds $\{\vec{b}_k\}_{k=1}^L$ of compatible sizes and a  vector $\vec{a}_L$, the DCNN can be defined by
\begin{equation}\label{DCNN}
    h_L(x)=  \hat{a}_L\cdot \vec{h}_L(x),
\end{equation}
where $\vec{h}_0(x)=x$,
\begin{equation}\label{DCNN-1}
      \vec{h}_k(x)=\sigma(\vec{w}_k\odot\vec{h}_{k-1}(x)+\vec{b}_k), \qquad k=1,\dots,L,
\end{equation}
 $\sigma$ acts on vectors componentwise and $\odot$ denotes either $*$ in \eqref{convolution} or $\star$ in \eqref{convolution1}.   Due to (\ref{sparse-structure}), DCNNs can be regarded as special deep fully connected neural networks with specified sparse comvolutional structures imposed to weight matrices.  Figure 1 shows   structures of the mentioned two types of DCNNs.
 \begin{figure}[!t]
\begin{minipage}[b]{0.49\linewidth}
\centering
\includegraphics*[scale=0.18]{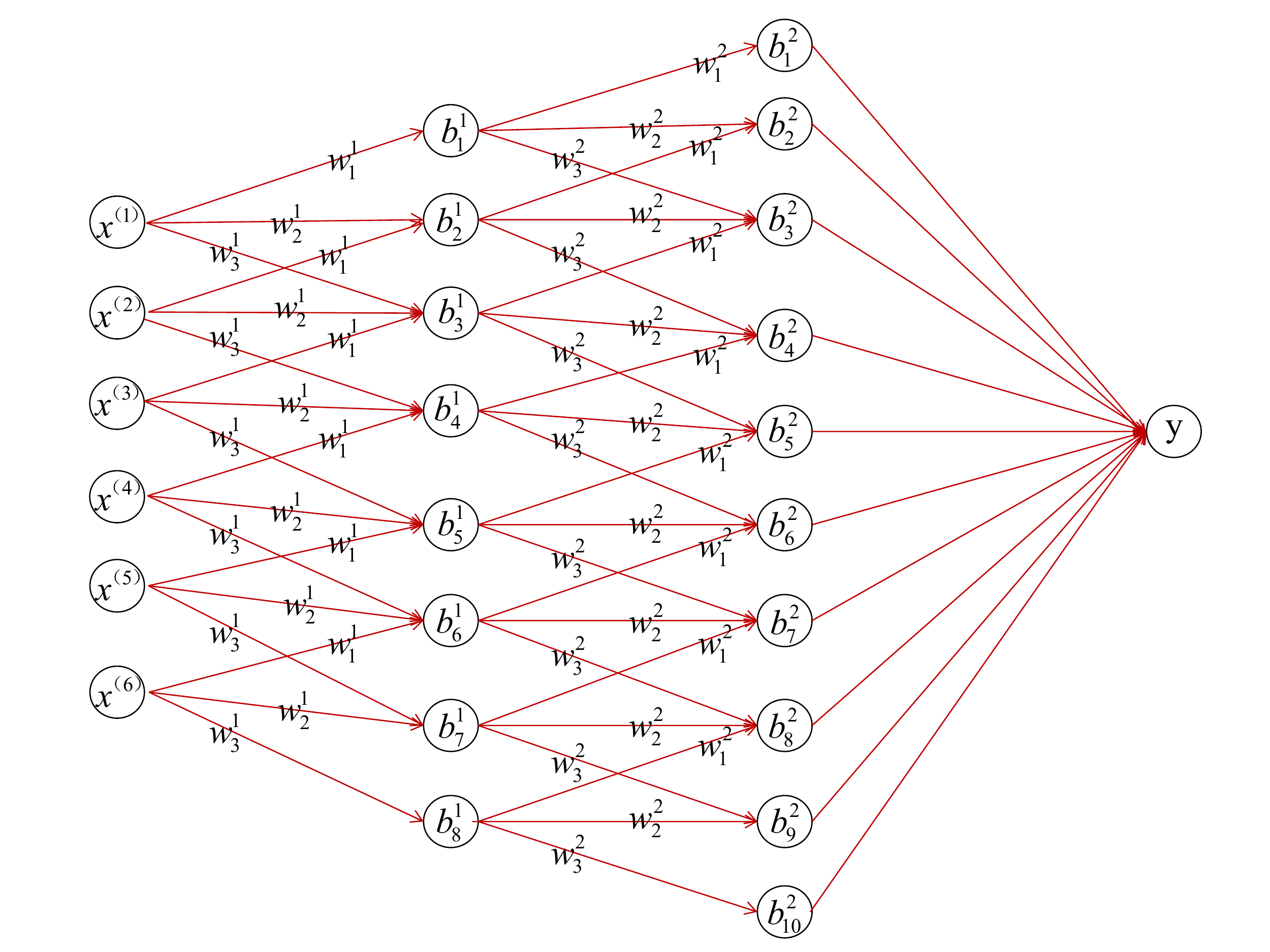}
\centerline{{\small (a) expansive DCNN}}
\end{minipage}
\hfill
\begin{minipage}[b]{0.49\linewidth}
\centering
\includegraphics*[scale=0.18]{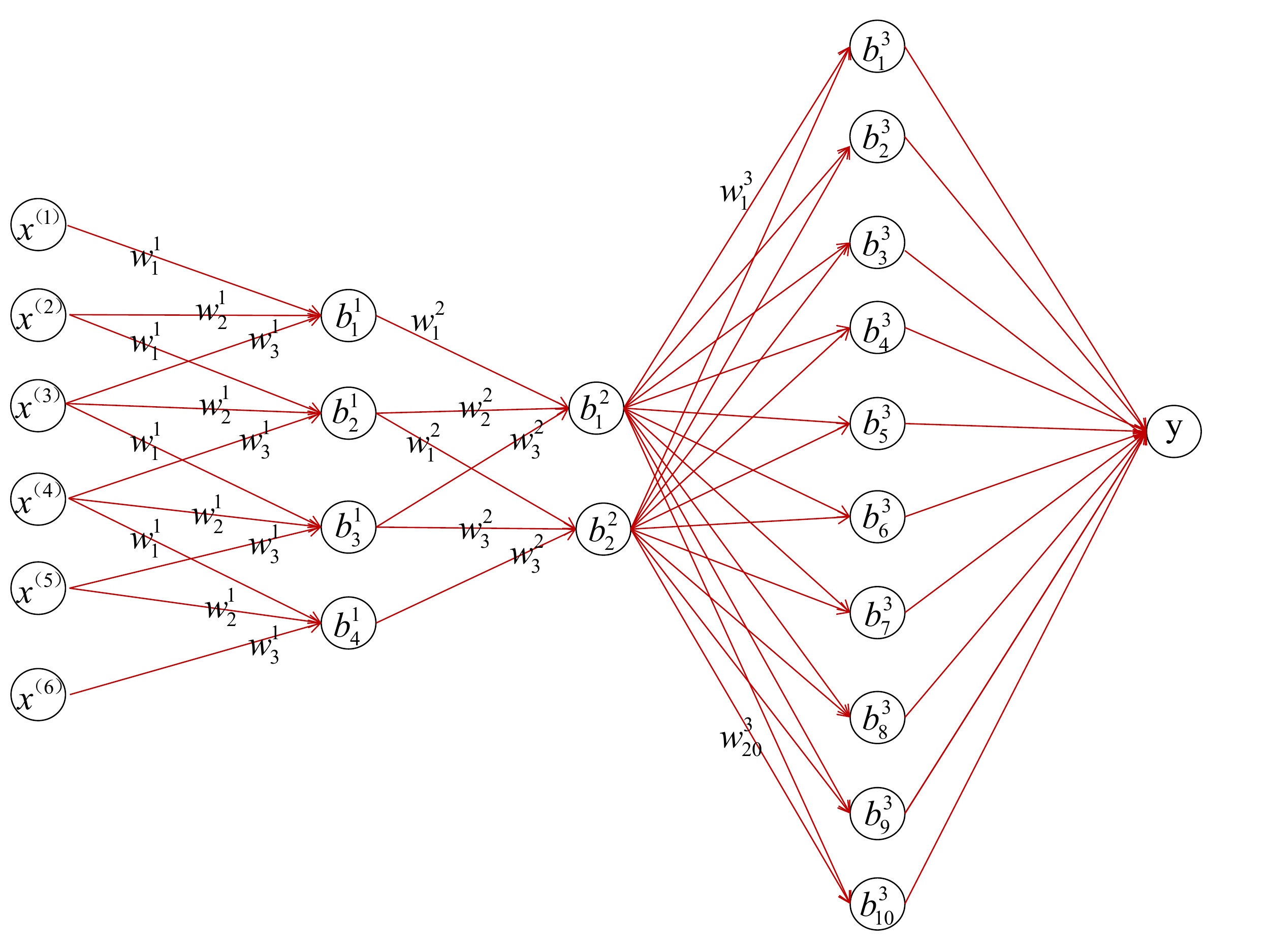}
\centerline{{\small (b) contracted DCNN}}
\end{minipage}
\hfill \caption{The left   shows the expansive DCNN with $s=2$ and $L=2$, while the right  exhibits a hybrid deep nets with  the contracted DCNN with $s=2$ and $L=2$ and a fully connected  neural network of width 10. } \label{fig:data}
\end{figure}

Practically, the contracting  DCNN (cDCNN) is more commonly used than the expansive DCNN (eDCNN) by utilizing  the convolutional layers to extract features of data. However, without fully connective layesr, it can be found in \cite{Hanin2019} that cDCNN is even not a   universal approximant, since the widths of all hidden layers are less than $d+1$. Differently,
the universal approximation property of eDCNNs has recently been verified in \cite{Zhou2020a}, showing that an eDCNN can approximate any continuous functions to an arbitrary accuracy, provided there are sufficiently many layers.
As discussed in \cite{Zhou2018,Zhou2020a,Zhou2020b}, a main advantage of  eDCNNs  is their perfect approximation capability in tackling high dimensional data,  though
the width and free parameters in eDCNNs increase  linearly with the depth. It should be mentioned that we can use either a threshold-sharing strategy in \cite{Zhou2020a} or a pooling-type down-sampling strategy \cite{Zhou2020b} to reduce them, while the universal approximation property   still holds.

The problem is, however, that the constructed network in \cite{Zhou2020a}   involves unbounded weights, which makes it   be not applicable for some learning purposes at the first glance (see the discussion in \cite[Sec. 2]{Oono2019} for example). To attack this problem, we use a tight  pseudo-dimension estimate of deep neural networks with continuous  piecewise polynomial activations derived in \cite{Bartlett2019} and a classical relation \cite{Haussler1992,Mendelson2003} between the pseudo-dimension and covering number, and succeed in deriving the universal consistency for implementing empirical risk minimization (ERM) on eDCNNs. We also conduct  a series of numerical experiments to verify our theoretical assertions and show the excellent learning performance of eDCNNs in real applications like human activity recognition and heartbeat classification.

\section{Universal Consistency of eDCNN}

In learning theory \cite{Cucker2007},  the samples in the data set $D:=\{z_i\}_{i=1}^m:=\{(x_i,y_i)\}_{i=1}^m$ are assumed to be drawn independently and identically from an unknown distribution $\rho$ on $Z:=\mathcal X\times\mathcal Y$, where $x_i\in\mathcal X\subseteq\mathbb R^d$, and $y_i\in\mathcal Y\subseteq R$. Throughout the paper, we assume $\mathcal X$ is a compact set. The aim is to learn a function $f_D$ based on $D$ to  minimize the generalization error
$$
        \mathcal E(f):=\int_{\mathcal Z}(f(x)-y)^2d\rho.
$$
Noting that   the regression function $f_\rho(x):=\int_{\mathcal Y}yd\rho(y|x)$ defined by means of the conditional distributions $\rho(\cdot|x)$ of $\rho$ minimizes the generalization error, our aim is then to find an estimator $f_D$ to minimize
\begin{equation}\label{equality}
      \mathcal E(f)-\mathcal E(f_\rho)=\|f-f_\rho\|_{L_{\rho_X}^2}^2,
\end{equation}
where $\rho_X$ is the marginal distribution of $\rho$ on ${\mathcal X}$.

  We build up the estimator via ERM:
\begin{equation}\label{ERM}
     f_{D,L,s}:={\arg\min}_{f\in\mathcal H_{L,s}}\mathcal E_D(f),
\end{equation}
where
$
     \mathcal E_D(f)=\frac1m\sum_{i=1}^m(f(x_i)-y_i)^2
$
denotes the empirical risk of $f$ and
\begin{equation}\label{hypothesis-space}
    \mathcal H_{L,s}:=\left\{c \cdot h_L(x): \vec{w}_k \ \mbox{is of length $s$},\ \vec{b}_k\in\mathbb R^{d+ks}, c \in\mathbb R^{d+Ls}\right\}
\end{equation}
be the set of all output functions produced by the eDCNN defined by (\ref{DCNN-1})  with $\odot=*$.
One of the most important properties that a learner should have  is
that, as the sample size grows, the deduced estimator
converges to the real relation between the input and output. This property, featured as
the strongly universal consistency \cite{Gyorfi2002},  can be defined as follows.

\begin{definition}\label{DEFINITION WEAKLY UNIVEAL}
 A
sequence of regression estimators $\{f_m\}_{m=1}^\infty$ is called
strongly universally consistent, if
$$
       \lim_{m\rightarrow\infty}\mathcal E(f_m)-\mathcal E(f_\rho)=0
$$
holds with probability one for all probability distributions $\rho$ satisfying
 $\int_{\mathcal Y}y^2d\rho(y|x) <\infty$.
\end{definition}

Our main result is the following theorem, which shows  that running ERM on eDCNN yields strongly universally consistent learners.
\begin{theorem}\label{Theorem:universal consistency}
Let $\theta\in(0,1/2)$ be an arbitrary  real number and $2\leq s\leq d$.  If   $L =L_m\rightarrow\infty$, $M =M_m\rightarrow\infty$, $M^2_mm^{-\theta}\rightarrow0$ and
\begin{equation}\label{condition of  universal}
           \frac{M_m^4L_m^2(L_m+d)\log L_m\log(M_m^2m)}{m^{1-2\theta}}\rightarrow0,
\end{equation}
then $\pi_{M_m}f_{D,L_m,s}$ is strongly universally consistent, where    $\pi_M t = \min\{M,|t|\}\cdot\mathrm{sgn}(t)$ is the well known truncation operator.
\end{theorem}

The proof of Theorem \ref{Theorem:universal consistency} can be found in Appendix A. It can be found in Theorem \ref{Theorem:universal consistency} that $M_m=\log m$ and $L_m=m^\alpha$ with $\alpha<1/3$ satisfy the assumptions and thus can yield strongly universally consistent estimator.
Generally speaking, boundedness of free parameters  play a crucial role in the classical literature of learning with neural networks \cite{Bartlett}. In particular, without any restrictions on free parameters,  it can be found in \cite{Maiorov1999,Maiorov1999a} that there exists a bounded sigmoid  function such that the pseudo-dimension of a deep net with this activation function, two hidden layers  and $\mathcal O(d)$ free parameters  is infinite, which implies that it is impossible to derive universal consistency for running ERM on such deep nets.  On the contrary, with a controllable magnitude of free parameters, the universal consistency holds for deep nets with an arbitrary bounded sigmoid activation function \cite{Anthony2009}. The main breakthrough in Theorem \ref{Theorem:universal consistency} is that without any restrictions on  free parameters,
implementing ERM on DCNNs also yields universally consistent estimators. The main reason for this breakthrough is the piecewise linear property of ReLU, which is crucial to derive tight pseudo-dimension estimates for eDCNNs \cite{Bartlett2019}.

The universal consistency in Theorem \ref{Theorem:universal consistency} demonstrates the versatility of eDCNNs for different learning tasks, which is totally different from cDCNN that requires different fully connected layers for different learning tasks. This phenomenon is also verified by our real data experiments, where eDCNNs with the same structure are adaptive for different data but cDCNNs need different fully connected layers to enhance their learning performance (See Appendix B).

\section{Numerical Experiments}\label{Sec:experiments}
In this section, we shall  illustrate the versatility of  eDCNNs through several simulated data and real data examples.
\subsection{Simulated data examples}
 We consider the following regression model
 \begin{equation}\label{sim equa}
 y=\dfrac{\sin(\|x\|_2)}{\|x\|_2} + \varepsilon, \end{equation}
 for generating training data, where $x$ is a random vector with entries uniformly distributed in $[-10, 10]$, and $\varepsilon$ is a random Gaussian noise with mean $0$ and variance $0.01$. To verify our theoretical assertion, we mainly consider three cases of the dimension of $x$, that is, the dimension $d$ varies in $\{30, 100, 1000\}$. Then by using (\ref{sim equa}), we generate the training data sets with the number $m$ varying in $\{100, 300, 500, 1000, 2000, \cdots, 9000, 10000\}$ for each $d$. For the network structure, we fix the filter length $s$ as $2$, and the number of network layers as $ L=\text{ceil}(\sqrt[4]{m}) $ that is consistent with the assumption of our theorem,  where $ \text{ceil}(\cdot) $ returns the value of a number rounded upwards to the nearest integer. To evaluate the prediction performance of the trained network, we further generate the test data sets in the same way as the training data, except that they are computed without noise, that is, $y_{test}=\dfrac{\sin(\|x_{test}\|_2)}{\|x_{test}\|_2}$. The number of test data is chosen as 2000 for $d=30,100$, and $10000$ for $d=1000$, respectively.

\begin{figure*}[htbp]
	\centering	
	\subfigure[]{
		\includegraphics[width=0.313\linewidth]{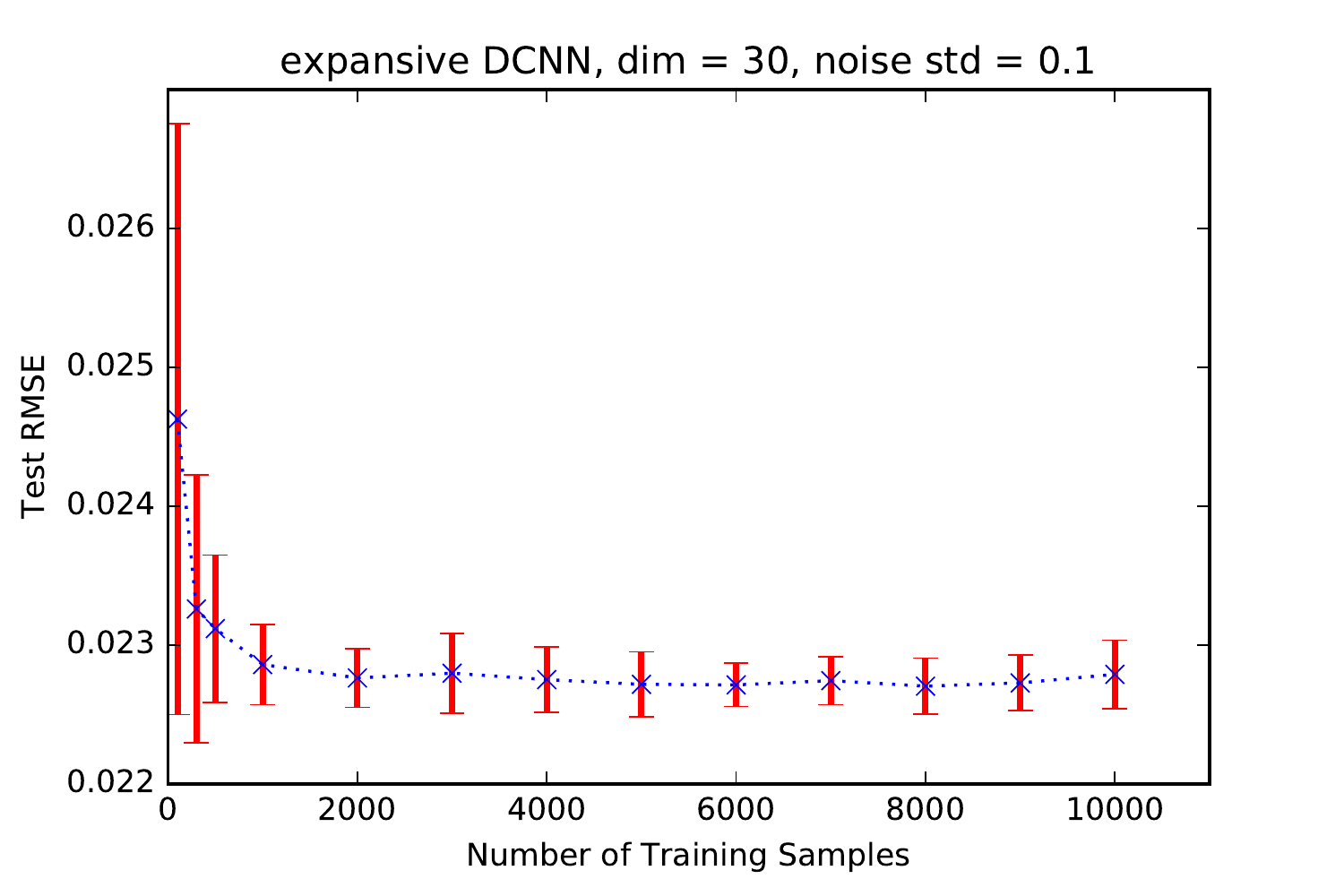}
	}
	\subfigure[]{
		\includegraphics[width=0.313\linewidth]{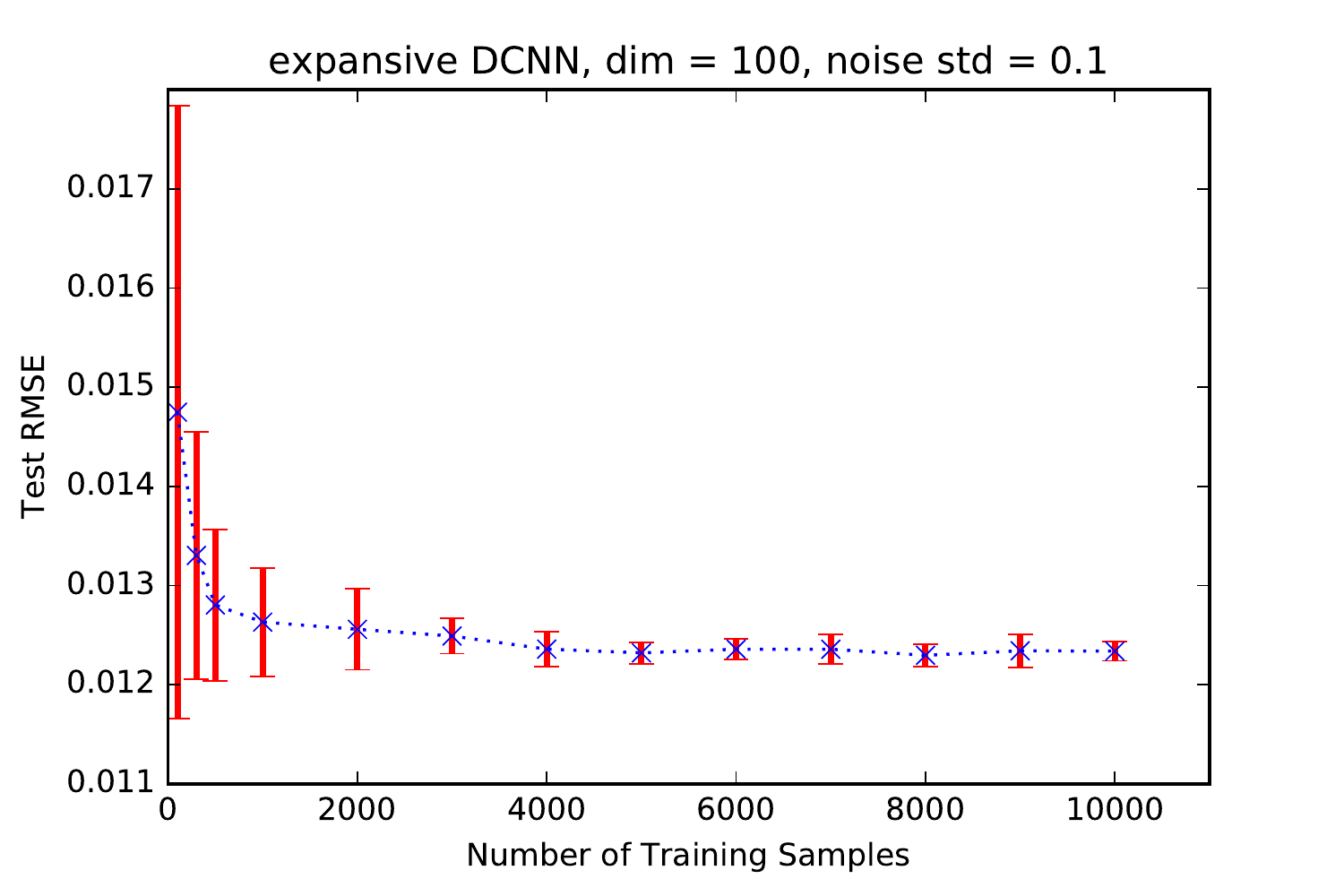}
	}
	\subfigure[]{
		\includegraphics[width=0.313\linewidth]{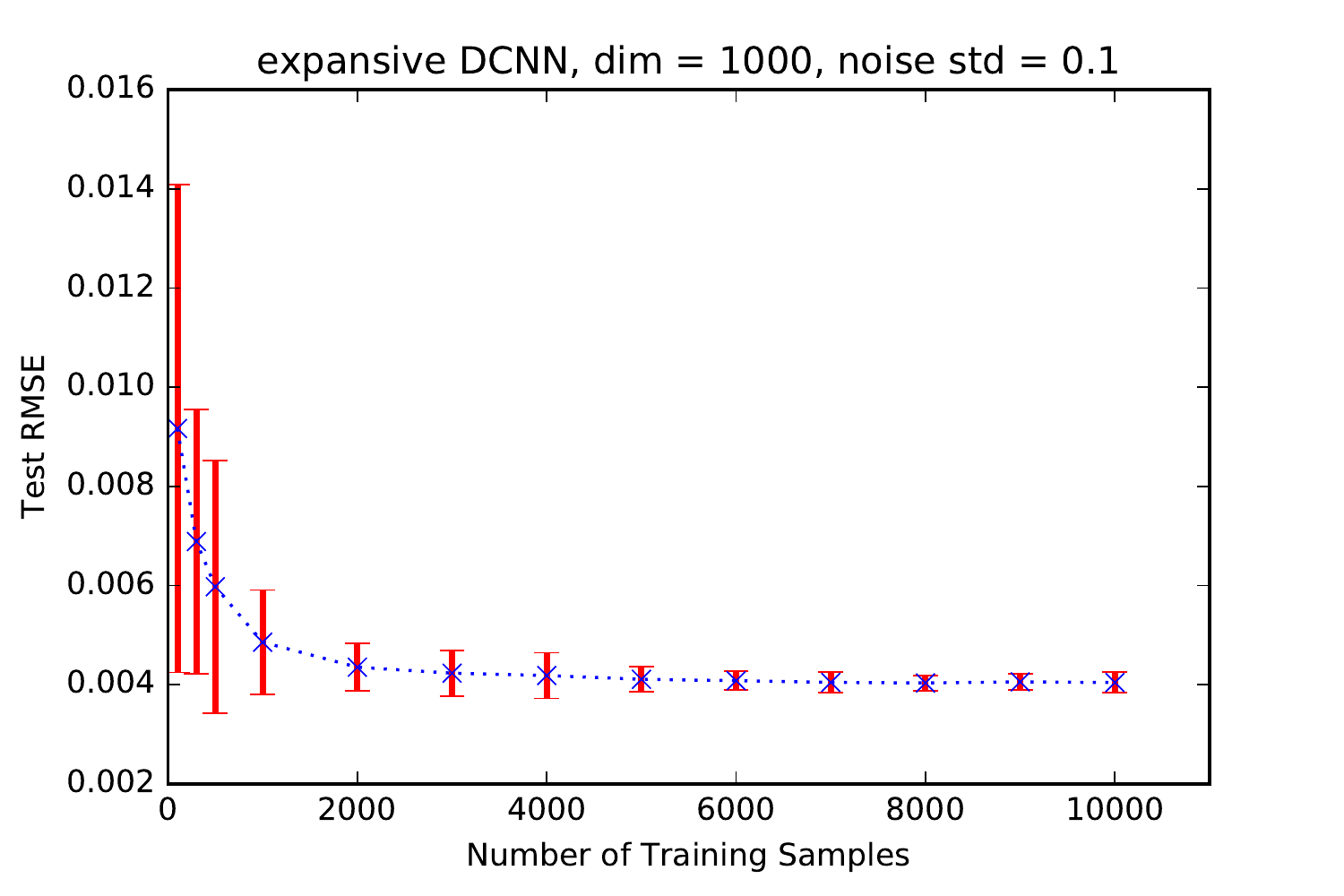}
	}
	\caption{The predication results of simulated data sets. (a) the error bar of case $ d=30 $; (b) the error bar of case $ d=100 $; (c) the error bar of case $ d=1000 $.}
	\label{simulation_results}
	\end{figure*}

 Fig. \ref{simulation_results} depicts the average results over 20 independent trials in terms of RMSE (root-mean-square error). It is not hard to observe from this figure that, for all the three cases, the test RMSE gradually decreases and then reaches a stable manner as the number of training samples $m$ grows.  This conforms Theorem \ref{Theorem:universal consistency}, since  $\mathcal E(f_{D,L,s})- \mathcal E(f_\rho)\rightarrow 0$ implies $\mathcal E(f_{D,L,s})\rightarrow  \mathcal E(f_\rho)$.

\subsection{Real data examples}
We now apply the proposed eDCNNs to deal with two real-world applications.

1. {\textit{Human Activity Recognition.}} In this application, we would like to recognize the type of movement (walking, running, jogging, etc.)  based on a given set of accelerometer data from a mobile device carried around a person's waist.  The data set considered here is the WISDM data set firstly released by \cite{Kwapisz2010} and includes 1098207 samples with 6 categories. Different from the methods used in \cite{Kwapisz2010}, we consider the traditional 1D cDCNN equipped with some fully connected layers as the baseline method for comparison. 
For the network structure of our eDCNN, we fix the filter length as $s=9$, and the number of network layers varies in $\{2, \ldots, 8\}$. The detailed architectures of the proposed eDCNN and the baseline network, and data descriptions can be found in Appendix B.

2.  {\textit{ECG Heartbeat Classification.}} An ECG is a 1D signal that is the result of recording the electrical activity of the heart using an electrode. It is a very useful tool that cardiologists use to diagnose heart anomalies and diseases. The data sets considered here are the MIT-BIH Arrhythmia Database and the PTB Diagnostic ECG Database that were preprocessed by \cite{Kachuee2018}. The MIT-BIH Arrhythmia data set includes 109446 samples with 5 categories, and the PTB Diagnostic ECG Database includes 14552 samples with 2 categories. We also compare the eDCNN with a traditional 1D CNN whose network architecture can be found in Appendix B.  In this application, we design our eDCNN in the same way as the Human Activity Recognition application, except that  the filter length is changed to $s=19$. One can find more details about the description and structures for this real application in Appendix B.

 Fig. \ref{real_results} shows the comparison results of the proposed eDCNN over the traditional 1D cDCNN in terms of the misclassification rate on the test data. It is easy to see that, as for the WISDM data set, eDCNN could obtain a slight better result than the 1D cDCNN for some choices of network layers;  as for the MIT-BIH Arrhythmia Database and the PTB Diagnostic ECG Database, eDCNN gives stably comparable result as the 1D cDCNN for a wide range of network layers. Considering the good theoretical guarantees and simple  structures (see Appendix B) of  eDCNN, we would prefer it over the traditional 1D cDCNN in practice.

\begin{figure*}[htbp]
	\centering	
	\subfigure[]{
		\includegraphics[width=0.313\linewidth]{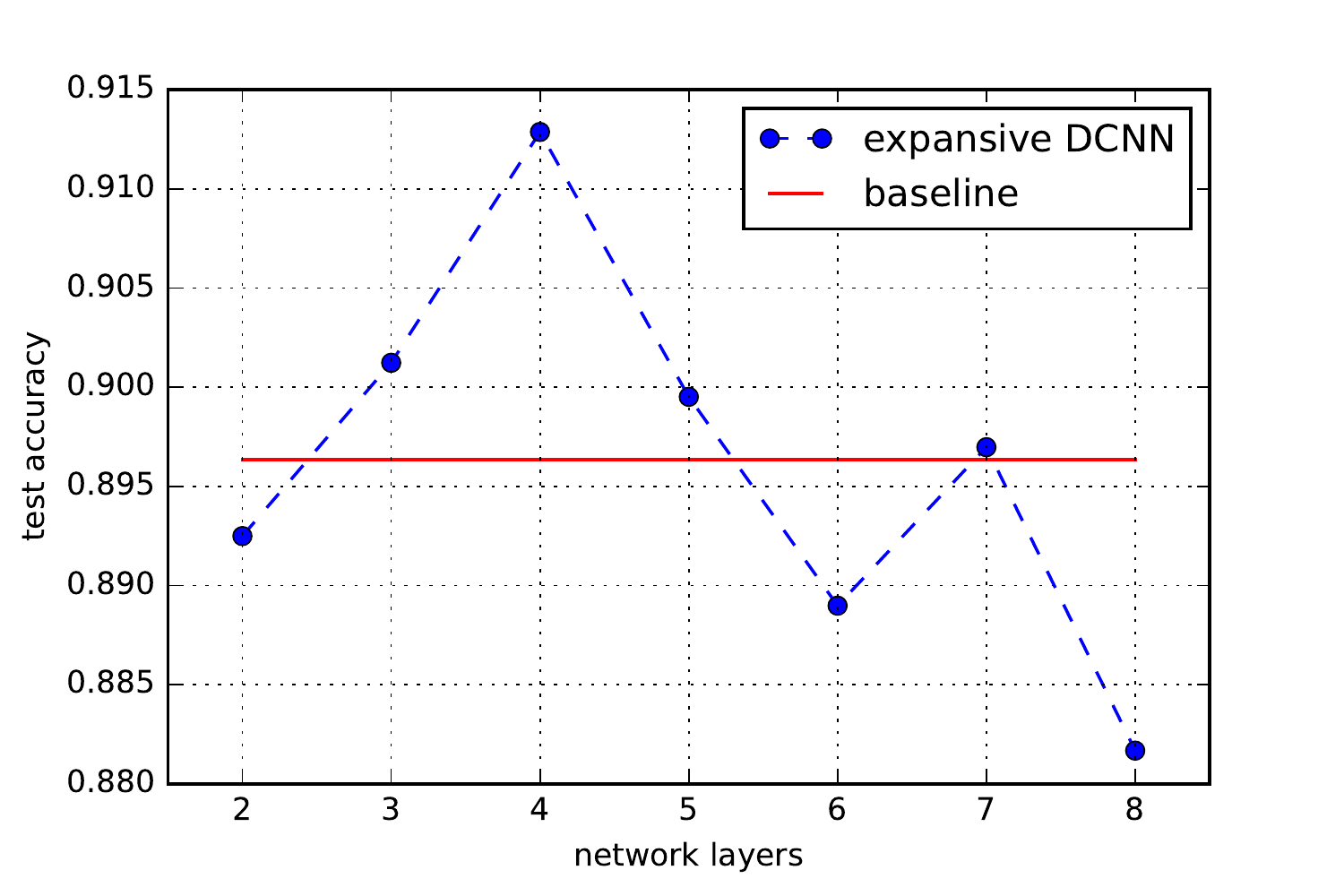}
	}
	\subfigure[]{
		\includegraphics[width=0.313\linewidth]{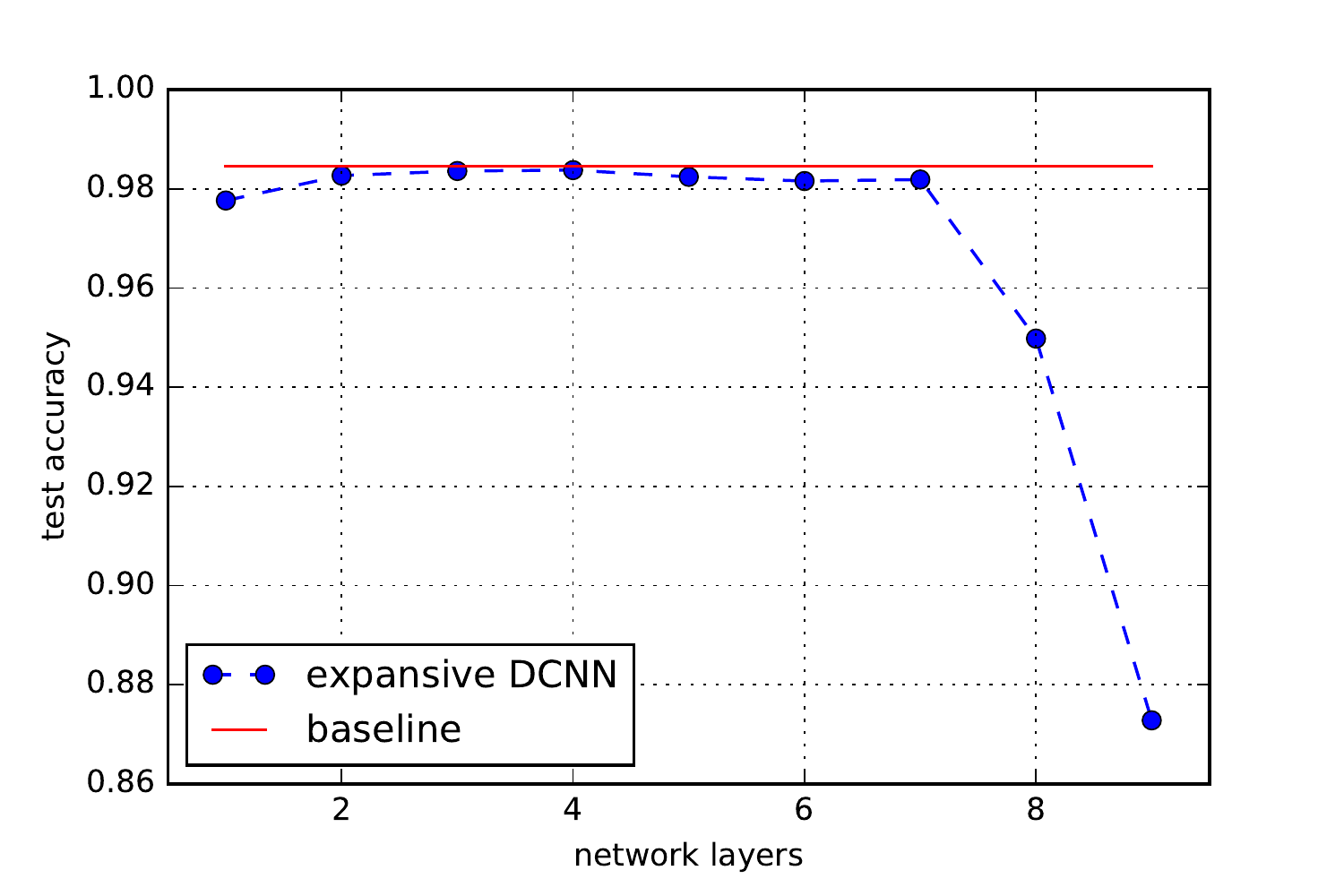}
	}
	\subfigure[]{
		\includegraphics[width=0.313\linewidth]{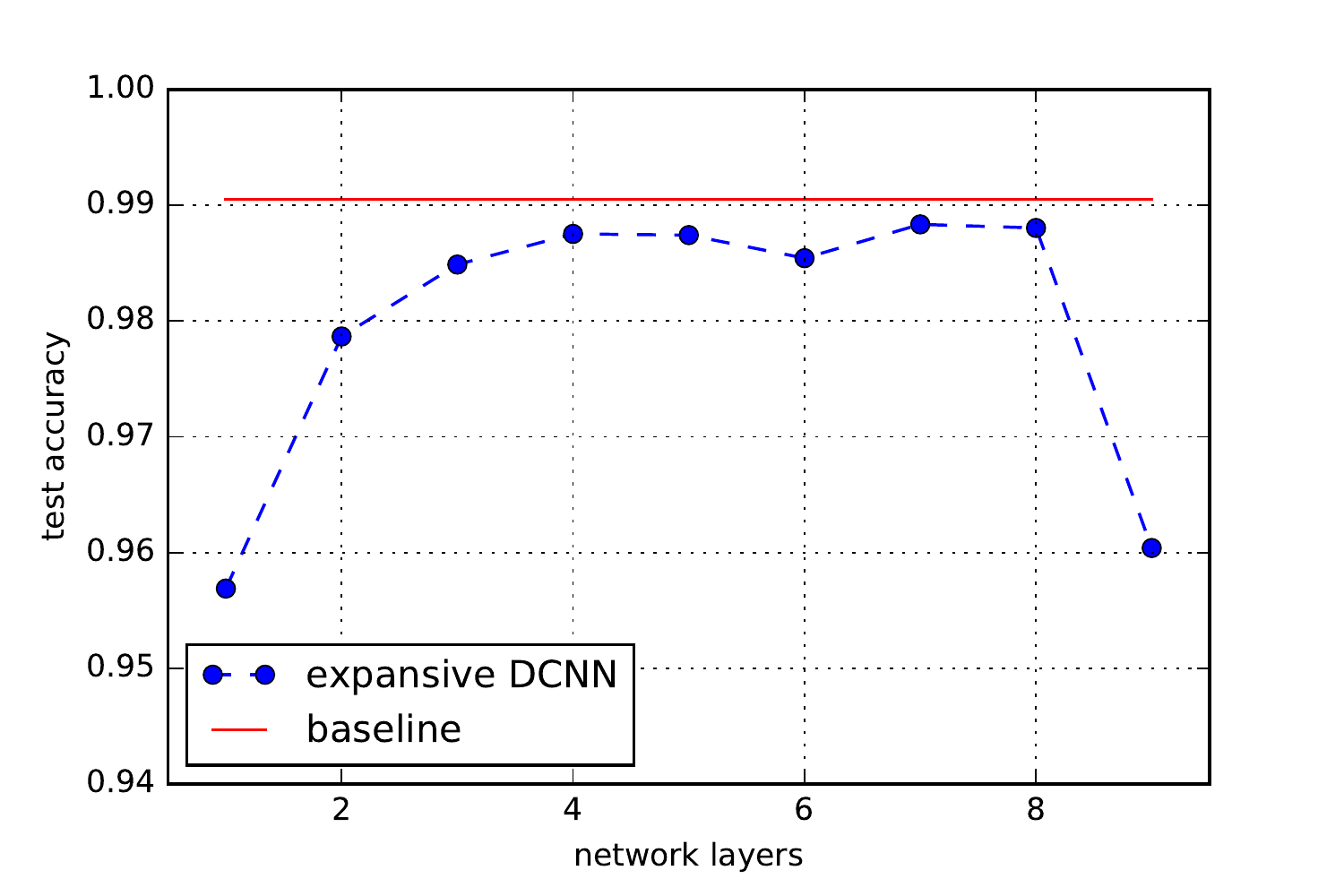}
	}
	\caption{The comparison results of real data sets. (a) the WISDM dataset; (b) the MIT-BIH Arrhythmia Database; (c) the PTB Diagnostic ECG Database.}
	\label{real_results}
	\end{figure*}

\section*{Acknowledge} {The work  of S. B. Lin is supported partially by   the  National Key R\&D Program of China (No.2020YFA0713900) and
   the National Natural Science Foundation of China
(No.618761332).  The work of Y. Wang is supported partially by the National Natural Science Foundation of China
(No.11971374).
 The work of D. X. Zhou  is supported partially by the Research Grants Council of Hong Kong [Project \# CityU 11307319], Hong Kong Institute for Data Science, and National Science Foundation of China [Project No. 12061160462]. This paper was written when the last author visited SAMSI/Duke during his sabbatical leave. He would like to express his gratitude to their hospitality and financial support.}

\begin{appendix}

\subsection{ Proof of Theorem \ref{Theorem:universal consistency}}

We divide our proof into three parts: capacity estimate, error analysis for bounded samples and universal consistency.

\subsubsection{Capacity estimate}
Let $\nu$ be a probability measure on $\mathcal X$. For a  function $f:\mathcal X\rightarrow\mathbb R$, set
$\|f\|_{L^p(\nu)}:=\left\{\int_{\mathcal X}|f(x)|^pd\nu\right\}^p$. Denote by $L^p (\nu)$ the set of all functions satisfying $\|f\|_{L^p(\nu)}<\infty$.
For $\mathcal V\subset L^p (\nu)$, denote by
$\mathcal N(\epsilon,\mathcal V,\|\cdot\|_{L^p(\nu)})$ the covering number \cite[Def. 9.3]{Gyorfi2002} of $\mathcal V$ in $L^p(\nu)$, which is
 the number of
elements in a least $\varepsilon$-net of $\mathcal V$ with respect to $\|\cdot\|_{L^p(\nu)}$. In particular, denote by $\mathcal N_p(\epsilon,\mathcal V,x_1^m):=\mathcal N(\epsilon,\mathcal V,\|\cdot\|_{L^p(\nu_m)})$ with $\nu_m$ the empirical measure with respect to $x_1^m=(x_1,\dots,x_m)\in\mathcal X^m$.  Define further $\mathcal M(\epsilon,\mathcal V,\|\cdot\|_{L^p(\nu_n)})$ to be the $\varepsilon$-packing number of $\mathcal V$ with respect to $\|\cdot\|_{L^p(\nu)}$, i.e. $\mathcal M(\epsilon,\mathcal V,\|\cdot\|_{L^p(\nu_n)})$ is the largest integer $N$ such that a subset $\{g_1,\dots,g_N\}$ of $\mathcal V$ satisfies $\|g_j-g_k\|_{L^p(\nu)}\geq \varepsilon$ for $1\leq j<k\leq N$. For the sake of brevity, we also denote $\mathcal M_p(\epsilon,\mathcal V,x_1^m):=\mathcal M(\epsilon,\mathcal V,\|\cdot\|_{L^p(\nu_m)})$ with respect to $x_1^m=(x_1,\dots,x_m)\in\mathcal X^m$. The following lemma found in \cite[Lemma 9.2]{Gyorfi2002} presents a relation between $\varepsilon$-covering numbers and $\varepsilon$-packing numbers.

\begin{lemma}\label{Lemma:pack-v.s.-cover}
Let $\mathcal V$ be a class of functions on $\mathcal X$  and let $\nu$ be a probability measure on $\mathcal X$, $p\geq 1$ and $\varepsilon>0$. Then
$$
   \mathcal M(2\varepsilon,\mathcal V,\|\cdot\|_{L^p(\nu)})
   \leq
   \mathcal N(\varepsilon,\mathcal V,\|\cdot\|_{L^p(\nu)})
   \leq
   \mathcal M(2\varepsilon,\mathcal V,\|\cdot\|_{L^p(\nu)}).
$$
In particular,
$$
    \mathcal M_p(2\epsilon,\mathcal V,x_1^m)
    \leq \mathcal N_p(\epsilon,\mathcal V,x_1^m)
    \leq
    \mathcal M_p(\epsilon,\mathcal V,x_1^m).
$$
\end{lemma}

Denote further by $Pdim(\mathcal V)$ the pseudo-dimension \cite[Chap. 14]{Anthony2009} of $\mathcal V$, which is the largest integer $\ell$  for which there exists $(\xi_1,\dots,\xi_m,\eta_1,\dots,\eta_m)\in\mathcal X^m\times\mathbb R^m$ such that for any $(a_1,\dots,a_\ell)\in\{0,1\}^{\ell}$ there exists some $v\in\mathcal V$ such that
$$
    \forall\ i:\quad v(\xi_i)>\eta_i\Leftrightarrow a_i=1.
$$
The following lemma that can be found in \cite[Theorem 6]{Haussler1992} (see also \cite[Theorem 1]{Mendelson2003}) presents a close relation between $\varepsilon$-packing numbers and pseudo-dimensions.

\begin{lemma}\label{Lemma:pseud-v.s.-packing}
Let $R>0$ and $\mathcal V_R$ be a set of functions from $\mathcal X$ to $[-R,R]$. Let $\nu$ be a probability measure. Then for any $\varepsilon\in(0,R]$, there holds
$$
    \mathcal M(2\varepsilon,\mathcal V_R,\|\cdot\|_{L^1(\nu)})\leq 2\left(\frac{2eR}{\varepsilon}\ln\frac{2eR}{\varepsilon} \right)^{Pdim(\mathcal V_R)}.
$$
\end{lemma}

From (\ref{DCNN-1}), there are $s+1$ tunable weights and $d+ks$ tunable thresholds in the $k$-th layers for $k=1,\dots,L-1$. Noting  additional $d+Ls$ tunable outer weights in the $L$-th layer, there are totally
\begin{equation}\label{free-parameter}
         n_{L,s}:=(s+1)L+d+Ls+\sum_{k=1}^L (d+ks)
\end{equation}
free parameters paved on
\begin{equation}\label{neurons}
    d_{L,s}:= 1+d+\sum_{k=1}^L(d+ks)
\end{equation}
neurons in the eDCNN.

Our main tool is a tight pseudo-dimension estimate for deep nets with piecewise linear activation.   In fact,
combining  \cite[Theorem 7]{Bartlett2019} and \cite[Theorem 14.1]{Anthony2009}, we can get  the following pseudo-dimension estimate for the eDCNN without any restrictions on the magnitudes of free parameters.
\begin{lemma}\label{Lemma:Pseudo-for-DCNN}
There exists an absolute constant $C_0$ such that
\begin{equation}\label{bound-for-pseudo}
    Pdim(\mathcal H_{L,s})\leq C_0Ln_{L,s}\log d_{L,s},
\end{equation}
where $n_{L,s}$ and $d_{L,s}$ are given in \eqref{free-parameter} and \eqref{neurons} respectively.
\end{lemma}

Our aim is to use the above three lemmas to derive a tight bound of the covering numbers of eDCNNs.
%
%
For $M>0$, define
\begin{equation}\label{trunction-hypothesis-space}
     \pi_M\mathcal H_{L,s}:=\{\pi_Mf: f\in\mathcal H_{L,s}\}.
\end{equation}
Since $Pdim(\pi_M\mathcal H_{L,s})\leq Pdim(\mathcal H_{L,s})$ \cite[p. 297]{Maiorov1999}, it follows from Lemma \ref{Lemma:Pseudo-for-DCNN} that
$$
   Pdim(\pi_M\mathcal H_{L,s})\leq C_0Ln_{L,s}\log d_{L,s}.
$$
Plugging the above estimate into Lemma \ref{Lemma:pseud-v.s.-packing}, we then have
$$
    \mathcal M(2\varepsilon,\pi_M\mathcal H_{L,s},\|\cdot\|_{L^1(\nu)})\leq 2\left(\frac{2eM}{\varepsilon}\right)^{2C_0Ln_{L,s}\log d_{L,s}}.
$$
Then it follows from Lemma \ref{Lemma:pack-v.s.-cover} with $\nu=\nu_m$ with respect to an arbitrary $x_1^m\in\mathcal X^m$ the following covering number estimates for the eDCNN without any restrictions to the magnitudes of parameters.
\begin{lemma}\label{Lemma:covering-number}
For any $0<\varepsilon\leq M$, there holds
$$
   \log_2\sup_{x_1^m\in\mathcal X^m}\mathcal N_1(\epsilon,\pi_M\mathcal H_{L,s},x_1^m)\leq c^* L^2(Ls+d)\log (L(s+d)) \log\frac{M}\epsilon,
$$
where $c^*$ is an absolute constant.
\end{lemma}

 \subsubsection{Error analysis for bounded samples}
Write $y_M=\pi_My$ and $y_{i,M}=\pi_My_i$. Define
$$
   \mathcal E_{\pi_M}(f)=\int_{\mathcal Z}(f(x)-y_M)^2d\rho,
$$
and
$$
     \mathcal E_{\pi_M,D}(f)=\frac1m\sum_{i=1}^m(f(x_i)-y_{i,M})^2.
$$
In this part, we aim at bounding $\mathcal E_{\pi_M}(\pi_Mf_{D,L,s})-\mathcal E_{\pi_M,D}(\pi_Mf_{D,L,s})$. Our  tool is the following concentration inequality, which can be easily deduced from
\cite[Theorem 11.4]{Gyorfi2002}.

\begin{lemma}\label{Lemma:CONCENTRATION INEQUALITY 1}
Assume $|y|\leq B$ and $B\geq 1$. Let $\mathcal F$ be a set of functions $f:\mathcal X\rightarrow\mathbb R$ satisfying $|f(x)|\leq B.$ Then for each $m\geq1$, with confidence at least
$$
      1-14\max_{x_1^m\in\mathcal X^m}\mathcal N_1\left(\frac{\beta\epsilon}{20B},\mathcal F,x_1^m\right)\exp\left(-\frac{\epsilon^2(1-\epsilon)\alpha m}{214(1+\epsilon)B^4}\right),
$$
there holds
\begin{eqnarray*}
     &&\sup_{f\in\mathcal F} \{\mathcal E(f)-\mathcal E(f_\rho)-( \mathcal E_D(f)-\mathcal E_D(f_\rho))\}\\
     &\leq&
     \epsilon(\alpha+\beta+ \mathcal E(f)-\mathcal E(f_\rho) ),
\end{eqnarray*}
where $\alpha,\beta>0$ and $0<\epsilon\leq1/2$.
\end{lemma}
Based on Lemma \ref{Lemma:CONCENTRATION INEQUALITY 1} and Lemma \ref{Lemma:covering-number}, we can derive the following lemma.

\begin{lemma}\label{Lemma:error-bounded}
If $M^2_mm^{-\theta}\rightarrow0$ and (\ref{condition of  universal}) holds for some $\theta\in(0,1/2)$, then
$$
     \lim_{m\rightarrow\infty}
     \mathcal E_{\pi_M}(\pi_Mf_{D,L,s})-\mathcal E_{\pi_M,D}(\pi_Mf_{D,L,s})=0
$$
holds almost surely.
\end{lemma}

\begin{IEEEproof} Since $|\pi_Mf_{D,L,s}(x)|,|y_M|,|y_{i,M}|\leq M$, we have
$$
   |\mathcal E_{\pi_M}(\pi_Mf_{D,L,s})-\mathcal E_{\pi_M,D}(\pi_Mf_{D,L,s})|
   \leq 8M^2.
$$
Then it follows from Lemma \ref{Lemma:CONCENTRATION INEQUALITY 1} with $\alpha=\beta=1$ and $\epsilon=m^{-\theta}$ that
with confidence at least
$$
      1-14\max_{x_1^m\in\mathcal X^m}\mathcal N_1\left(\frac{1}{20Mm^{\theta}},\pi_M\mathcal H_{L,s},x_1^m\right)\exp\left(-\frac{m^{1-2\theta}}{428M^4}\right),
$$
there holds
\begin{eqnarray*}
     \mathcal E_{\pi_M}(\pi_Mf_{D,L,s})-\mathcal E_{\pi_M,D}(\pi_Mf_{D,L,s})
     \leq 8M^2m^{-\theta}.
\end{eqnarray*}
Due to Lemma \ref{Lemma:covering-number}, we have
\begin{eqnarray*}
     &&
   \max_{x_1^m\in\mathcal X^m}\mathcal N_1\left(\frac{1}{20Mm^{\theta}},\pi_M\mathcal H_{L,s},x_1^m\right)\exp\left(-\frac{m^{1-2\theta}}{428M^4}\right)\\
   &\leq &
   \exp\left( c^*\log(20 M^2m^\theta) L^2 (d+sL)\log (L(s+d))-\frac{m^{1-2\theta}}{428M^4}\right).
\end{eqnarray*}
 Noting (\ref{condition of  universal}), we obtain
$$
    \lim_{m\rightarrow\infty}\max_{x_1^m\in\mathcal X^m} \mathcal N_1\left(\frac{1}{20M_mm^{\theta}},\pi_{M_m}\mathcal H_{L_m,s},x_1^m\right)\exp\left(-\frac{m^{1-2\theta}}{428M_m^4}\right)
    =0.
$$
Thus, as $m\rightarrow\infty$,
\begin{eqnarray*}
     \mathcal E_{\pi_{M_m}}(\pi_{M_m}f_{D,L_m,s})-\mathcal E_{\pi_{M_m},D}(\pi_{M_m}f_{D,L_m,s})
     \leq 8M^2m^{-\theta}\rightarrow0
\end{eqnarray*}
holds almost surely. This completes the proof of Lemma \ref{Lemma:error-bounded}.
\end{IEEEproof}
 \subsubsection{Universal consistency}
 Our final tool is the universality of eDCNNs, which was proved in \cite[Theorem 1]{Zhou2020a}.
\begin{lemma}\label{Lemma:approx}
 Let $2\leq s\leq d$. For any compact subset $\mathcal X$ of $\mathbb R^d$ and any $f\in C(\mathcal X)$,  there exists an $h_{L,s}\in\mathcal H_{L,s}$ such that
\begin{equation}\label{app}
    \lim_{L\rightarrow+\infty}\|f-h_{L,s}\|_{C(\mathcal X)}=0.
\end{equation}
\end{lemma}

Now we are in a position to prove Theorem \ref{Theorem:universal consistency}.

\begin{IEEEproof}[Proof of Theorem \ref{Theorem:universal consistency}]
 Since $\mathbf E\{y^2\}<\infty$, we have
$f_\rho\in L^2 ({\rho_X})$.
It follows from Lemma \ref{Lemma:approx} that for any $\varepsilon>0$, there exists some
 $g_\varepsilon\in \mathcal  H_{L_\varepsilon,s}$  with sufficiently large $L_\varepsilon$ such that
\begin{equation}\label{Jackson type}
            \|f_\rho-g_\varepsilon\|_{L^2 ({\rho_X})}^2\leq
           \varepsilon.
\end{equation}
The triangle inequality then yields
\begin{eqnarray*}
              &&\mathcal E(\pi_Mf_{D,L,s})-\mathcal E(f_\rho)\\
              &\leq&
              \mathcal E(\pi_Mf_{D,L,s})-(1+\varepsilon)\mathcal E_{\pi_M}(\pi_Mf_{D,L,s}) \\
              &+&
              (1+\varepsilon) (\mathcal E_{\pi_M}(\pi_Mf_{D,L,s}) -
             \mathcal E_{\pi_M,D}(\pi_Mf_{D,L,s}))\\
             &+&
             (1+\varepsilon) (\mathcal E_{\pi_M,D}(\pi_Mf_{D,L,s}) -
             \mathcal E_{\pi_M,D}(f_{D,L,s}))\\
             &+&
             (1+\varepsilon) \mathcal E_{\pi_M,D}( f_{D,L,s})-(1+\varepsilon)^2\mathcal E_{D}( f_{D,L,s}) \\
             &+&
             (1+\varepsilon)^2( \mathcal E_{D}( f_{D,L,s})-
             \mathcal E_{D}(g_\varepsilon)) \\
             &+&
             (1+\varepsilon)^2(\mathcal E_{D}(g_\varepsilon)-\mathcal E(g_\varepsilon))\\
             &+&
             (1+\varepsilon)^2(\mathcal E(g_\varepsilon)-\mathcal E(f_\rho))\\
             &+&
             ((1 +\varepsilon)^2-1)\mathcal E(f_\rho)\\
             &=:&
             \sum_{\ell=1}^8B_\ell.
\end{eqnarray*}
To deduce the strongly universal consistency, we should bound $B_\ell$,
$\ell=1,\dots,8$, in probability, respectively. As
\begin{equation}\label{abinequality}
         (a+b)^2\leq(1+\varepsilon)a^2+(1+1/\varepsilon)b^2 \ \  \mbox{for} \ a,b>0,
\end{equation}
we have
\begin{eqnarray*}
       B_1
       &=&
       \int_{\mathcal Z}|\pi_Mf_{D,L,s}(x)-y_M+y_M-y|^2d\rho\\
              &-&(1+\varepsilon)\int_{\mathcal Z}|\pi_Mf_{D,L,s}(x)-y_M|^2d\rho \\
              &\leq&
              (1+1/\varepsilon)\int_Z|y-y_M|^2d\rho.
\end{eqnarray*}
Since $M=M_m\rightarrow \infty$ as $m\rightarrow\infty$, we obtain
$$
          B_1\rightarrow 0\ \ \mbox{when} \ m\rightarrow
           \infty.
$$
From Lemma \ref{Lemma:error-bounded}, (\ref{condition of  universal}) and $M^2_mm^{-\theta}\rightarrow0$, it follows that
$$
          B_2\rightarrow 0\ \ \mbox{when} \ m\rightarrow
           \infty
$$
holds almost surely.
%
%
 The definition of the truncation
operator yields
$$
            \frac1m\sum_{i=1}^m|\pi_Mf_{D,L,s}(x_i)-y_{i,M}|^2
            -\frac1m\sum_{i=1}^m| {f}_{D,L,s}(x_i)-y_{i,M}|^2\leq 0.
$$
Therefore, we have
$$
             B_{3}\leq 0.
$$
According to the strong law of large numbers and
(\ref{abinequality}), we get
$$
         B_{4}\leq(1+\varepsilon)(1+1/\varepsilon)
         \frac1m\sum_{i=1}^m|y_i-y_{i,M}|^2\rightarrow
          (1+\varepsilon)(1+1/\varepsilon)\int_Z |y-y_{M}|^2d\rho
$$
as $m\rightarrow \infty$ almost surely. Therefore,
$M_m\rightarrow\infty$ and the definition of $y_M$ yield
$$
           B_{4}\rightarrow 0.
$$
 Due to (\ref{ERM}),  we obtain
$$
           B_5=(1+\varepsilon)^2\left(\frac1m\sum_{i=1}^m| {f}_{D,L}(x_i)-y_i|^2-\frac1m\sum_{i=1}^m|g_\varepsilon(x_i)-y_i|^2\right) \leq
           0.
$$
By the strong law of large numbers again, we have almost surely
$$
          B_{6}\rightarrow0, \ \ \ (\hbox{as} \ m\rightarrow\infty).
$$
To bound $ B_{7}$,  we note from (\ref{equality})
$$
          B_{7}=(1+\varepsilon)^2\|g_\varepsilon-f_\rho\|_\rho^2.
$$
The above equality together with (\ref{Jackson type}) and $L_m\rightarrow\infty$ yields
$$
          B_{7}\leq (1+\varepsilon)^2\varepsilon.
$$
Noting
$$
           (1+\varepsilon)^2-1=\varepsilon(\varepsilon+2),
$$
we have
$$
           B_{8}\leq
           \varepsilon(\varepsilon+2)\int_Z|f_\rho(x)-y|^2d\rho.
$$
Using all the above assertions, we can concludes that
$$
         \lim\sup_{m\rightarrow\infty} \mathcal E(\pi_Mf_{D,L,s})-\mathcal E(f_\rho) \leq
         (1+\varepsilon)^2\varepsilon+
         \varepsilon(\varepsilon+2)\int_Z|f_\rho(x)-y|^2d\rho
$$
holds almost surely. This proves Theorem \ref{Theorem:universal consistency}  by setting
$\varepsilon\rightarrow0.$
\end{IEEEproof}

\begin{figure}[!t]
	\centering	
	\subfigure[]{
		\raisebox{0.5\height}{\includegraphics[width=0.193\linewidth]{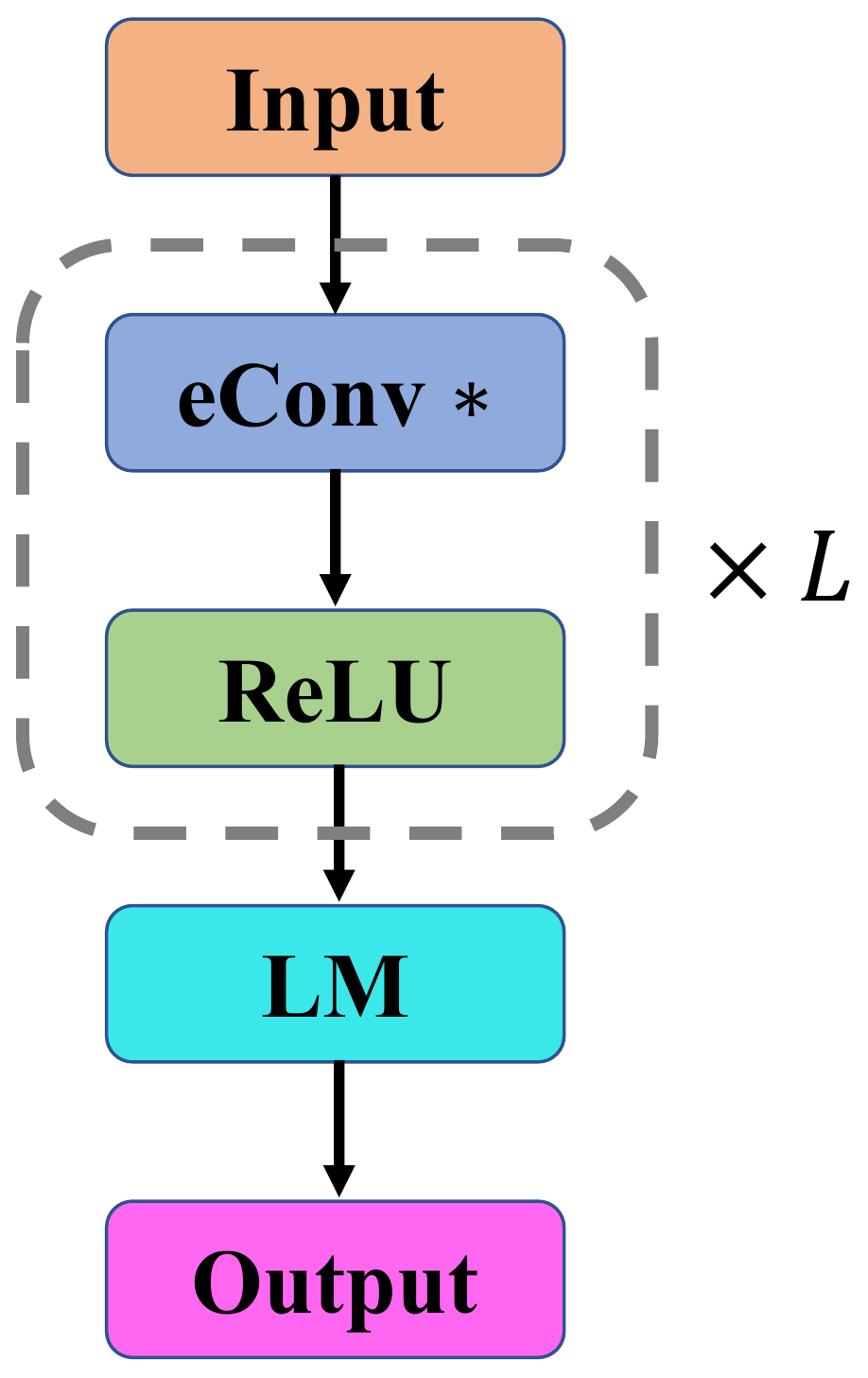}}
	}
		\hspace{-0.4cm}
	\subfigure[]{
		\raisebox{0.2\height}{\includegraphics[width=0.199\linewidth]{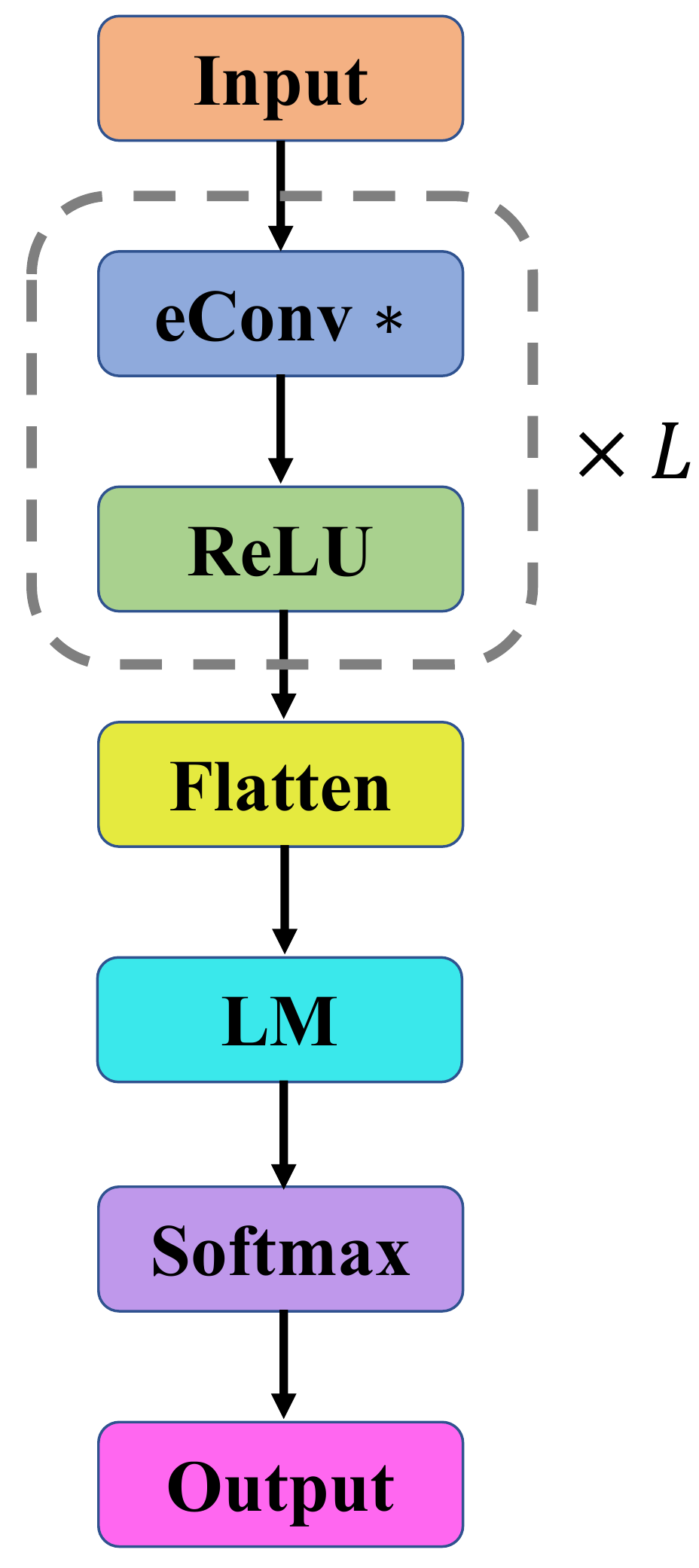}}
	}
		\hspace{-0.4cm}
	\subfigure[]{
		\includegraphics[width=0.216\linewidth]{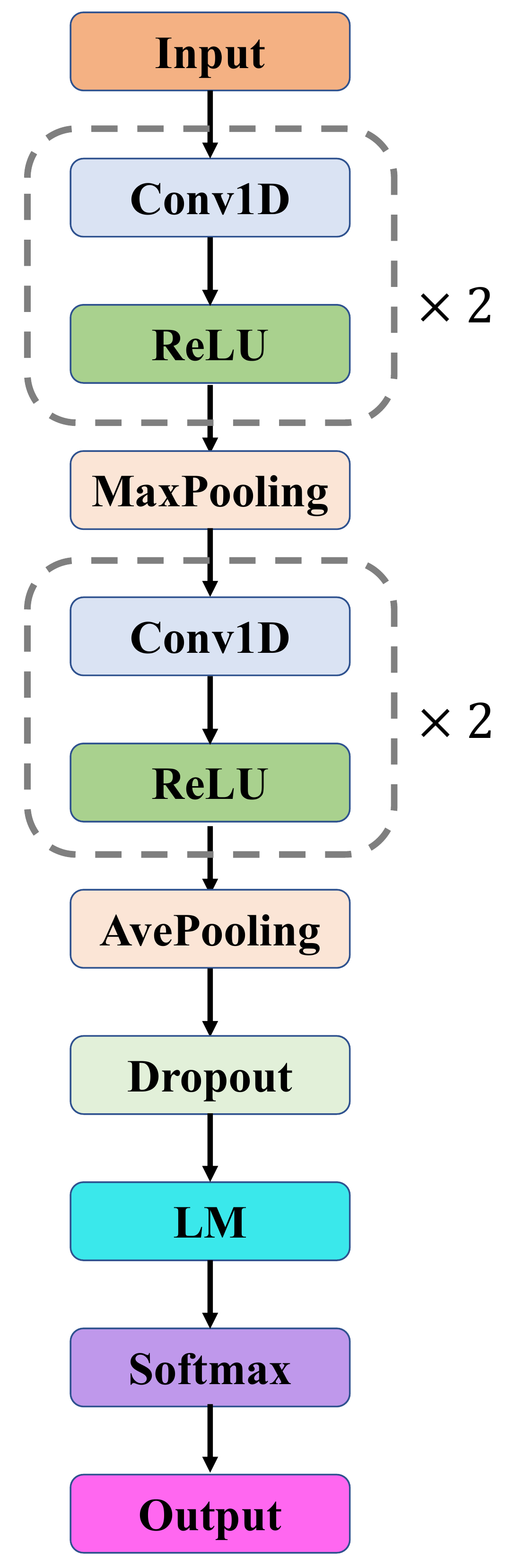}
	}
		\hspace{-0.4cm}
	\subfigure[]{
		\raisebox{0.02\height}{\includegraphics[width=0.294\linewidth]{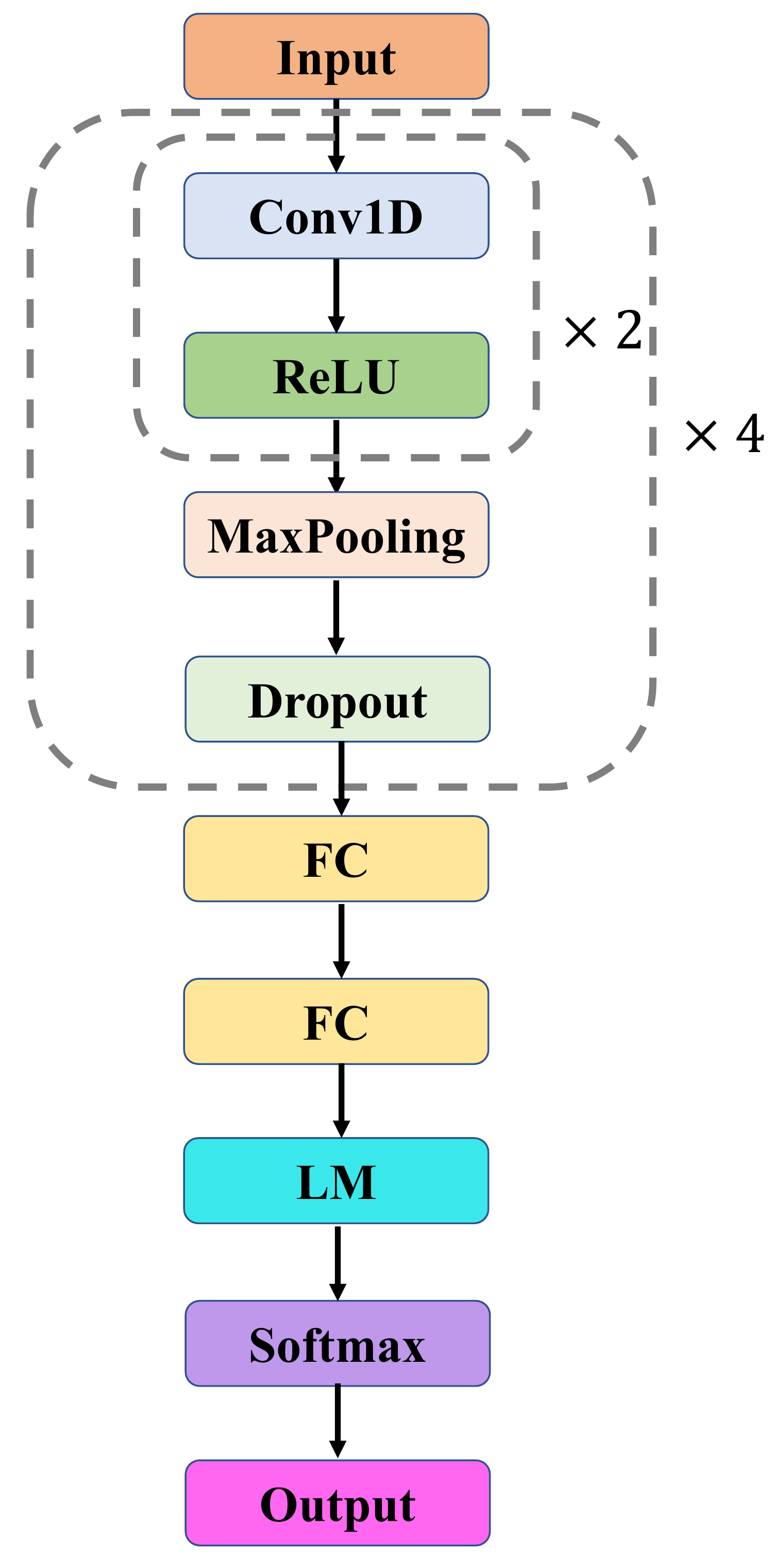}}
	}
	\caption{The detailed network architectures of our eDCNN and baseline methods used in the simulated and real data experiments, where the ``LM" module means linear mapping and the ``FC" module is a fully connected layer. (a) eDCNN for simulated data examples; (b) eDCNN for real data examples; (c) baseline network for the Human Activity Recognition task; (d) baseline network for the ECG Heartbeat Classification task.}
	\label{network structure}
\end{figure}

\subsection{ Description and structures for real data sets}
In this part, we introduce some auxiliary information for our real data experiments.

\subsubsection{Detailed description of implementing real data sets}
We shall  provide more details on how to implement the three real data sets using the proposed eDCNNs.

1. \textit{WISDM dataset. } This dataset is a collection of accelerometer data taken from a smartphone that various people carried with them while conducting six different exercises, i.e., Downstairs, Jogging, Sitting, Standing, Upstairs, and Walking. For each exercise, the accelerations with respect to the x, y, and z axes were measured and captured with a timestamp and person ID.

We first split all the samples based on the user IDs, that is, users with ID 1 to 28 are used for training the model and users with ID greater than 28 are consider as the test set. We then reshape the data by considering 80 time periods as one record with the accelerations with respect to the x, y, and z axes. Therefore, the input of our network is a matrix of size $80\times3$ and the output of our network is a vector with length 6. To train the network, we choose the cross entropy function as the loss function, and the adam as the optimizer. Then the batch size and the maximal number of epochs are set as 400 and 50, respectively. Similar to other traditional CNNs, the early stopping strategy is also used for training out network.

2. \textit{MIT-BIH Arrhythmia database.} The MIT-BIH Arrhythmia Database contains 48 half-hour excerpts of two-channel ambulatory ECG recordings, obtained from 47 subjects studied by the BIH Arrhythmia Laboratory between 1975 and 1979. It was preprocessed by ~\cite{Kachuee2018} based on the beat extraction method described in III.A of the paper, and thus the data set has totally 109446 samples corresponding to five categories.

To test the performance of our network, we randomly choose 80\% samples as the training set and the remaining 20\% as the test set.  Since each sample has 187 attributes, the input of our network is a vector with length 187 and the output of our network is a vector with length 5. Similar to before, the cross entropy function is chosen as the loss function, the adam is consider as the optimizer and the early stopping strategy is used to train our network. Then the batch size and the maximal number of epochs are set as 32 and 100, respectively.

3. \textit{PTB Diagnostic ECG Database.} The database contains 549 records from 290 subjects (aged 17 to 87, mean 57.2; 209 men, mean age 55.5, and 81 women, mean age 61.6; ages were not recorded for 1 female and 14 male subjects). Each subject is represented by one to five records.  By using the beat extraction method in \cite{Kachuee2018}, this data set is transformed to a new one that has totally 14552 samples corresponding to two categories.  The implementation details of this data is the same as the MIT-BIH Arrhythmia database, except that the network's output is a vector with length two.

\subsubsection{Architectures of the implemented DCNNs}
Fig. \ref{network structure} plots the detailed network architectures of our eDCNNs and baseline networks used in the simulated and real data experiments. It is easy to see that the network structures of the proposed eDCNNs are simpler and cleaner than the baseline networks.

\end{appendix}


\begin{thebibliography}{99}

\bibitem{Allen-Zhu2019}
Z.~Allen-Zhu, Y.~Li, and Z.~Song.  A convergence theory for deep learning via
  over-parameterization. ICML, 2019.


  \bibitem{Anthony2009}
M. Anthony and P. L. Bartlett. Neural Network Learning: Theoretical Foundations. Cambridge University Press, 2009.


%
%
%
\bibitem{Bartlett}
P. Bartlet. The sample complexity of pattern classification with
neural networks: the size of the weights is more important than the
size of the networks. IEEE Trans. Inf. Theory, 44: 525-536, 1998.


\bibitem{Bartlett2019}
P. L. Bartlett, N. Harvey, C.  Liaw, and A. Mehrabian. Nearly-tight VC-dimension and Pseudodimension Bounds for Piecewise Linear Neural Networks. J. Mach. Learn. Res.,  20(63): 1-17, 2019.


\bibitem{Bengio2013}
Y. Bengio, A. Courville, and P. Vincent. Representation learning: A
review and new perspectives. IEEE Trans. Pattern Anal. Mach. Intel.,
 3: 1798-1828, 2013.





%








\bibitem{Chui2020}
C. K. Chui, S. B. Lin, B. Zhang, and D. X. Zhou.
 Realization of spatial sparseness by deep ReLU nets with massive data.
\emph{IEEE Transactions on Neural Networks and Learning Systems}, In Press, 2020.





\bibitem{Cucker2007}
F. Cucker and D. X. Zhou.  Learning Theory: an Approximation Theory
Viewpoint.   Cambridge University Press, Cambridge, 2007.
%
%





%


\bibitem{Goodfellow2016}
I. Goodfellow, Y. Bengio, and A. Courville. {Deep Learning}. MIT
Press, 2016.


\bibitem{Guo2020}
Z. C. Guo, S. Lei, and S. B. Lin. Realizing data features by deep nets. IEEE Trans. Neural Netw. Learn. Syst., In press, 2020.

\bibitem{Gyorfi2002}
L. Gy\"{o}rfy, M. Kohler, A. Krzyzak, and H. Walk. A Distribution-Free
Theory of Nonparametric Regression. Springer, Berlin, 2002.

\bibitem{Hanin2019}
B. Hanin. Universal function approximation by deep neural nets with bounded width and relu activations. Mathematics, 7 (10): 992, 2019.

\bibitem{Haussler1992}
D. Haussler. Decision theoretic generalizations of the PAC model for
neural net and other learning applications. Inform. Comput., 100:
78-150, 1992.

\bibitem{He2016}
K. He, X. Zhang, S. Ren and J. Sun.
Deep residual learning for image recognition, CVPR, 2016.


\bibitem{Kachuee2018}
M. Kachuee, S. Fazeli, and M. Sarrafzadeh. ECG heartbeat classification: a deep transferable representation. In Proceedings of IEEE International Conference on Healthcare Informatics (ICHI), 2018.
%


%

\bibitem{Kohler2017}
M. Kohler and A. Krzyzak. Nonparametric regression based on
hierarchical interaction models.  IEEE Trans. Inf. Theory,   63:
 1620-1630, 2017.


\bibitem{Krizhevsky2012}
A.~Krizhevsky, I.~Sutskever, and G.~E. Hinton.  Imagenet classification with
  deep convolutional neural networks. NIPS, 1097--1105, 2012.

%




\bibitem{Kwapisz2010}
J. R. Kwapisz, G. M. Weiss, and S. A. Moore. Activity recognition using cell phone accelerometers. In Proceedings of the Fourth International Workshop on Knowledge Discovery from Sensor Data, 2010.
%


%
%


\bibitem{Lin2019}
S. B. Lin. Generalization and expressivity for deep nets. IEEE
Trans. Neural Netw. Learn. Syst.,   30:  1392-1406, 2019.



%
%
\bibitem{Maiorov1999}
V. Maiorov and J. Ratsaby. On the degree of approximation by
manifolds of finite pseudo-dimension. Constr. Approx., 15: 291-300,
1999.
%
%
\bibitem{Maiorov1999a}
V. Maiorov and A. Pinkus. Lower bounds for approximation by MLP
neural networks. Neurocomputing,   25: 81-91, 1999.
%

%
%



\bibitem{Mei2018}
S. Mei, A. Montanari, and P. M. Nguyen. A mean field view of the landscape of two-layer neural networks.
Proc. Nat. Acad. Sci. USA, 115 (33): E7665-E7671.


\bibitem{Mendelson2003}
S. Mendelson and R. Vershinin. Entropy and the combinatorial dimension.
Invent. Math., 125: 37-55, 2003.

%


\bibitem{Oono2019}
K. Oono and T. Suzuki. Approximation and non-parametric estimation of ResNet-type convolutional neural networks. ICML, 2019: 4922-4931.

%









\bibitem{Schmidt2020}
J. Schmidt-Hieber. Nonparametric regression using deep neural networks with ReLU activation function.
Ann. Statist., 48(4): 1875-1897, 2020.

\bibitem{Silver2016}
D.~Silver, A.~Huang, C.~J. Maddison, A.~Guez, L.~Sifre, G.~V.~D. Driessche,
  J.~Schrittwieser, I.~Antonoglou, V.~Panneershelvam, and M.~Lanctot.
   Mastering the game of go with deep neural networks and tree search.
  {Nature},  529(7587):  484--489, 2016.



%
%
%


%




\bibitem{Zhou2018}
D. X. Zhou. Deep distributed convolutional neural networks:
Universality. Anal. Appl.,  16: 895-919, 2018.

\bibitem{Zhou2020a}
D. X. Zhou. Universality of deep convolutional neural networks.
  Appl. Comput. Harmonic. Anal., 48: 784-794, 2020
%
%
\bibitem{Zhou2020b}
D. X. Zhou. Theory of deep convolutional neural networks: Downsampling. Neural Netw., 124: 319-327, 2020.



\end{thebibliography}
\end{document}